\documentclass[journal]{IEEEtai}
\usepackage{amsmath}
\usepackage{amssymb}
\usepackage{booktabs}
\usepackage[colorlinks,urlcolor=blue,linkcolor=blue,citecolor=blue]{hyperref}
\usepackage{multirow}
\usepackage{color,array}

\usepackage{graphicx}

\begin{document}

\title{Global-Local Image Perceptual Score (GLIPS): Evaluating Photorealistic Quality of AI-Generated Images}

\author{Memoona Aziz,~\IEEEmembership{Student Member,~IEEE}, Umair Rehman~\IEEEmembership{Member,~IEEE}, Muhammad Umair Danish ~\IEEEmembership{Student Member,~IEEE}, Katarina Grolinger, ~\IEEEmembership{Member,~IEEE}

\thanks{This work has been submitted to the IEEE for possible publication. Copyright may be transferred without notice, after which this version may no longer be accessible.}
\thanks{Submitted on: May 14, 2024}
        
\thanks{Memoona Aziz is with the Department of Computer Science, Western University, London, ON, Canada. E-mail: maziz86@uwo.ca}

\thanks{Umair Rehman is with the Department of Computer Science, Western University, London, ON, Canada. E-mail: urehman6@uwo.ca}

\thanks{Muhammad Umair Danish is with the Department of Electrical and Computer Engineering, Western University, London, ON, Canada. E-mail: mdanish3@uwo.ca}

\thanks{Katarina Grolinger is with the Department of Electrical and Computer Engineering, Western University, London, ON, Canada. E-mail: kgroling@uwo.ca}

%\thanks{Manuscript received Month DD, YYYY; revised Month DD, YYYY.}

}

\markboth{Under Review}
{Under Review}

\maketitle

\begin{abstract}
This paper introduces the Global-Local Image Perceptual Score (GLIPS), an image metric designed to assess the photorealistic image quality of AI-generated images with a high degree of alignment to human visual perception. Traditional metrics such as FID and KID scores do not align closely with human evaluations. The proposed metric incorporates advanced transformer-based attention mechanisms to assess local similarity and Maximum Mean Discrepancy (MMD) to evaluate global distributional similarity. To evaluate the performance of GLIPS, we conducted a human study on photorealistic image quality. Comprehensive tests across various generative models demonstrate that GLIPS consistently outperforms existing metrics like FID, SSIM, and MS-SSIM in terms of correlation with human scores. Additionally, we introduce the Interpolative Binning Scale (IBS), a refined scaling method that enhances the interpretability of metric scores by aligning them more closely with human evaluative standards. The proposed metric and scaling approach not only provide more reliable assessments of AI-generated images but also suggest pathways for future enhancements in image generation technologies.
\end{abstract}


\section{Introduction}
\IEEEPARstart{T}{he} widespread production of AI-generated images is reshaping the visual digital landscape. In 2023, over 15 billion images were generated using text-to-image generation algorithms by surpassing the combined photographic output of the past 150 years within just a single year since the technology's inception \cite{ej}. Since the launch of DALLE-2 \cite{openai_dalle_paper}, the daily production rates of AI-generated images have soared to approximately 34 million. with significant contributions from platforms such as Adobe Firefly, which reported 1 billion images created by users within three months of its launch \cite{ej}. Furthermore, the image-generating segment of the generative AI market was valued at \$299.2 million in 2023 and is predicted to reach \$917.4 million by 2030, marking a compound annual growth rate of 17.4\%. \cite{ej2}, \cite{talebi2018nima}. These figures highlight the importance of the increasing demand for photorealistic images by emphasizing the need for robust evaluation metrics that can align with human judgment in assessing their quality.

In parallel to the explosive growth, there exists an inherent challenge: the absence of metrics for assessing photorealistic image quality. While the surge in AI-generated visuals demonstrates technological prowess, it also amplifies the need for a metric that can objectively evaluate the photorealistic quality of these images against the standards of human perception. This problem persists across various domains, from entertainment to simulation training, where the distinction between synthetic and authentic visuals holds significant practical importance. For instance, in the healthcare domain, where AI-generated images are used for medical training, the ability to discern realistic imagery could be crucial to learning outcomes \cite{degardin2024fake}, \cite{wang2004image}.

In addressing the existing methods for photorealism and quality assessment, previous research has primarily divided the approaches into two distinct categories: pixel-based comparison and model-based evaluation \cite{wang2023high}. On the one hand, pixel-based metrics such as Structural Similarity Index Measure (SSIM) and Peak Signal-to-Noise Ratio (PSNR) have been traditionally employed to evaluate the structural similarity and noise discrepancies between AI-generated images and reference photographs  \cite{talebi2018learned}. On the other hand, model-based metrics leverage the sophistication of pre-trained neural networks' learned representations. The Fréchet Inception Distance (FID) \cite{heusel2017gans} and the Inception Score (IS) are prominent examples of model-based metrics \cite{NIPS2016_8a3363ab}. These metrics utilize neural network representations to discern photorealism, image quality, and diversity. Despite the advancements offered by these methods, challenges persist, particularly in aligning these metrics with human judgment. The reason behind this is that both camera-captured and AI-generated images have different underlying structures, which makes it challenging to compare them based on mathematical similarity.

Furthermore, challenges arise in fairly comparing image metric output and human scores by presenting a crucial challenge: the scaling of evaluation scores for a fair and accurate comparison. Traditional scaling methods, such as Z-score normalization or Min-Max scaling, are often employed to standardize the scores obtained from various metrics \cite{rehman2015display}. However, these techniques may inadvertently introduce bias by disproportionately amplifying the influence of certain metrics over others due to their naturally wider or narrower scoring ranges. For instance, a metric like PSNR, which typically operates on a larger numerical scale could unduly overshadow a metric like SSIM, which has a constrained score range. This incongruity presents a notable challenge and compels the adoption of a refined scaling methodology to guarantee an equitable, interpretable, and impartial comparative examination between computational assessments and human judgments.

\begin{figure*}[h!]
    \centering
    \includegraphics[width=0.99\textwidth]{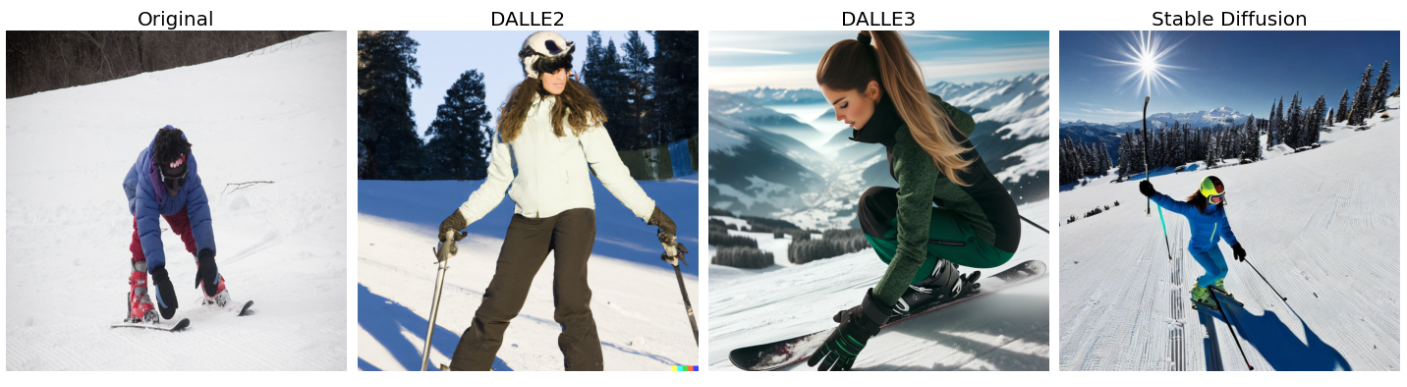} 
    \vspace{-8pt}
    \caption{Original and Images Generated by Models such as DALLE-2, DALLE-3, and Stable Diffusion. The caption given to the model while generating images is "A woman touching her skis going down a ski hill." The original image belongs to the MS-COCO dataset with ID: 000000080671.}
    \label{fig:image}
\end{figure*}

To overcome the inherent limitations of both traditional and model-based metrics, this paper introduces multiple contributions. First, we present a novel metric called the Global-Local Image Perceptual Score (GLIPS). GLIPS uniquely integrates vision transformer-based attention mechanisms to extract attention scores of important image patches. This accommodates the different structures of camera and AI-generated images. GLIPS also computes global distributional similarity between camera and AI-generated images to ensure complete essence of images have been captured. GLIPS specifically focuses on critical image patches by evaluating their overlap and direct differences between their representations, aiming to achieve high correlation with human perceptual judgments.

Second, we conducted a human study by designing and conducting a survey on Photorealistic Image Quality. The publicly available link to the human scores dataset can be found at: \href{https://github.com/udanish50/VisualVerity}{https://github.com/udanish50/VisualVerity}.

Third, to ensure fair comparison between human and metric scores, we introduce the Interpolative Binning Scale (IBS). IBS provides an interpretable transformation of both human scores and metric outputs into fair and understandable scores guided by human experts in the field of image processing.

This paper makes several significant contributions to the field of photorealistic image assessment:
\begin{enumerate}
    \item Design of the Global-Local Image Perceptual Score, which utilizes attention mechanisms to compute local patch similarity and the second computes the Maximum Mean Discrepancy of the entire image's deep features to assess global distributional similarity.
    \item Design of a novel scaling strategy employing the Likert scale and linear interpolation to ensure unbiased and interpretable comparison between human and metric scores.
    \item Design of Human Study on photorealistic image quality to assess and comprehend their correlation with image metric outputs.
    \item Comparative analysis and benchmarking of GLIPS against traditional photorealistic image quality metrics, including SSIM, Multi Scale-SSIM, FID and KID. 
\end{enumerate}

The GLIPS consists of two parts: the first extracts the attention scores of image patches to compute local patch similarity, and the second computes the MMD of the entire image's deep features to assess global distributional similarity. A regulatory parameter controls the balance between these two scores.

The remainder of the paper is organized as follows: Section \ref{sec:relwork} presents the related work, Section \ref{sec:ARPI} details the proposed Global-Local Image Perceptual Score and Interpolative Binning Scale, and Section \ref{sec:results} presents results and analysis. Finally, Section \ref{sec:conclusion} concludes the paper.

\section{Related Study}
\label{sec:relwork}
This section covers current metrics, image generative models, subjective assessments, and scaling strategies. 
\subsection{Image Metrics} 
Photorealistic image quality assessment is fundamentally categorized into pixel-based metrics and model-based distances \cite{wang2023high}. Pixel-based methods, such as SSIM evaluate structural resemblance by mimicking the human visual system's response to spatial variations in images \cite{talebi2018learned}. The Multi-Scale Structural Similarity Index Measure (MS-SSIM) introduces a more comprehensive evaluation by assessing image quality at various resolutions, thereby capturing both fine and coarse structural details. MS-SSIM is particularly effective in contexts where images may be viewed at multiple scales or when more sensitivity to changes in structural information is required \cite{wang2003multiscale}.

The PSNR calculates the squared discrepancies between a reference and a generated image, offering a straightforward measure of image degradation \cite{talebi2018learned}. Visual Information Fidelity (VIF) is another metric that quantifies the amount of visual information preserved in an image compared to a reference image, based on natural scene statistics and the characteristics of the human visual system \cite{1576816}. SSIMPLUS, introduced by Rehman et al. \cite{rehman2015display}, advances SSIM by integrating factors like human vision, display characteristics, and viewing conditions by enabling real-time perceptual quality predictions.

On the model-based front, the FID evaluates the statistical distance between feature vectors of real and generated images captured by the Inception model serving as an indicator of visual fidelity \cite{heusel2017gans}. The Inception Score (IS), developed by Salimans et al. \cite{NIPS2016_8a3363ab} quantifies image diversity and clarity using the predictive entropy of a classifier trained on diverse datasets like ImageNet. Kernel Inception Distance (KID), introduced by Binkowski et al. \cite{binkowski2018demystifying}, improves upon FID by employing a non-parametric approach using a polynomial kernel to compute the squared Maximum Mean Discrepancy between feature distributions. This method avoids assumptions about the distributional form of activations, better accommodating the non-negative nature of ReLU activations in deep networks.

The Learned Perceptual Image Patch Similarity (LPIPS) metric was developed by Zhang et al. \cite{zhang2018unreasonable}, offers a distinct approach by utilizing activations to directly measure perceptual similarity between images. This metric employs a weighted combination of activations across multiple layers of a pre-trained network, reflecting a more effective understanding of human visual perception compared to traditional pixel-based metrics.

The effectiveness of traditional metrics in assessing the photorealism and quality of images generated by advanced generative models is often limited. For example, image metrics generally only account for about 60\% of the variations in perceptual quality assessments made by human observers \cite{valdebenito2023exploration}. This highlights a significant gap where these metrics do not fully capture human visual perception, which is critical for evaluating the photorealism of AI-generated images. This discrepancy indicates a substantial opportunity for designing more advanced metrics that better align with human judgment and provide more accurate assessments.

\subsection{Scaling Strategies}
Scaling strategies in image quality assessment are critical for aligning metric scores with human perceptual judgments. Traditional methods like Z-standardization and Min-Max scaling can introduce biases by disproportionately emphasizing certain scores, such as giving undue weight to PSNR over SSIM. To address this, researchers have developed methods like linear regression to better align metric scores with mean opinion scores (MOS) from human evaluators, providing a more representative evaluation of image quality \cite{rehman2015display}.

Advanced models such as the Non-linear Receptive Field model offer a more nuanced approach by optimizing parameters to enhance correlation with human ratings, tailored to specific image datasets \cite{luna2023state}. Additionally, the Saliency-Guided Local Full-Reference Image Quality Assessment leverages visual saliency to weight local image quality scores, prioritizing regions that are more likely to capture human attention, thus potentially providing a more accurate reflection of human visual perception \cite{varga2022saliency}.

While these methods improve the relevance and accuracy of assessments, their complexity and black-box nature often make them less transparent and harder to trust in practical applications. This highlights a critical gap in the field: the need for an interpretable and fair scaling strategy that can be readily trusted by practitioners.

\vspace{-0.28 cm}
\subsection{Image Generative Models}
Recent advancements in image generative models (IGMs) have markedly transformed the field of image synthesis. OpenAI's DALL-E \cite{openai_dalle_paper}, a groundbreaking model, introduced the capability to generate detailed images from textual descriptions using a hybrid of Variational Autoencoders (VAEs) and Transformer \cite{openai_dalle_paper}. Its successors, DALL-E 2 and DALL-E 3, have built upon this foundation with improved resolution and fidelity further pushing the boundaries of text-to-image conversion. DALL-E 2 enhanced image detail at multiple scales, while DALL-E 3 integrates more advanced diffusion techniques to produce even more photorealistic outputs \cite{dosovitskiy2021image}, \cite{openai_dalle_paper}.

Other notable models include GLIDE (Guided Language to Image Diffusion for Generation and Editing), which also utilizes Transformer-based architectures and diffusion models to create high-quality images from text \cite{nichol2021glide}. Stable Diffusion stands out for its efficiency and accessibility by employing a Variational Autoencoder framework to enable rapid generation of images \cite{lee2023diffusion}. MidJourney and Google's Imagen have made significant contributions to the landscape of IGMs. MidJourney utilizes unique training approaches to create visually artistic interpretations from text prompts, while Imagen pushes the envelope on photorealism \cite{IMAGEN} \cite{ramesh2021zero, nichol2022glide}. 

\subsection{Subjective Assessment}
Human visual perception plays a crucial role in image comprehension, particularly in evaluating image quality and realism. Li et al. introduced a questionnaire for assessing the general quality and fidelity of generative images \cite{li2023agiqa}. Lu et al.  \cite{lu2024seeing} advanced the field by proposing HPBench which is specifically designed for the subjective evaluation of photorealism. Further exploring the intersection of visual perception and artificial intelligence, Ragot et al.  \cite{ragot2020ai} conducted a pivotal study that aimed to align painting captions with their corresponding visuals. Zhou et al.  \cite{zhou2023eyes} embarked on a multifaceted exploration into the subjective realms of beauty, liking, valence, and arousal by offering insights into emotional responses.  Treder et al. \cite{treder2022quality} focused on the subjective evaluation of image quality through a human-centred study, prioritizing human ratings to gauge image quality. In the realm of photorealistic image synthesis, Sarkar et al. \cite{sarkar2021humangan} introduced HumanGAN, a study that demonstrates the capabilities of generative models in crafting lifelike images. Complementing these efforts, Rassin et al.  \cite{rassin2022dalle} provided an in-depth analysis of text-image alignment within the DALL-E 2 framework, showcasing the potential for synergistic visual-textual narratives. Degardin et al.'s research on generative images depicting human actions \cite{degardin2024fake} and Sun et al.'s gender bias assessment in DALL-E 2 \cite{sun2024smiling} further highlight the complexity and ethical considerations inherent in generative AI.

\section{Proposed Methods}
This section will explain the Global-Local Image Perceptual Score (GLIPS) and the Interpolative Binning Scale method in detail to provide a comprehensive understanding of their theoretical framework.
\label{sec:ARPI}

\subsection{Global-Local Image Perceptual Score (GLIPS)}
The GLIPS consists of two parts: the first extracts the attention scores of image patches to compute local patch similarity, and the second computes the Maximum Mean Discrepancy of the entire image's deep features to assess global distributional similarity. A regulatory parameter controls the balance between these two scores.

\vspace{+2 mm}
\noindent \subsubsection{Salient Patch Extraction Algorithm}
The major challenge in assessing camera-captured and AI-generated images arises from their differing structures. This difference is evident in Figure \ref{fig:image}, where images generated from the same caption, 'A Woman touching her skis going down a ski hill,' show variability. There are two common elements in these images: the woman and the ski hill. However, the position, color, and standing angle of the woman vary significantly between images. This variability poses a substantial challenge for standard metrics to accurately compute their similarity. To address this issue, we leverage a pre-trained Vision Transformer (ViT) \cite{han2022survey}, which enables us to obtain attention scores for important image patches. This method is likely to assign higher scores to patches with more significant intensity values, ultimately providing a more appropriate measure of similarity for AI-generated images.

Let's consider two images, $I_o$ for the original camera-generated image and $I_g$ for the AI-generated image. The salient patch extraction algorithm aims to identify and extract the most relevant patches from input images based on ViT attention mechanisms. Given input images, the algorithm operates directly on the outputs of the Multi-Head self-attention of the Vision Transformer as follows:

\noindent Initially, the images \(I_o\) and \(I_g\) are processed through the ViT, which includes computing attention maps:
\begin{equation}
A_o = \text{ViT}(I_o)
\end{equation}
\begin{equation}
A_g = \text{ViT}(I_g)
\end{equation}

\noindent Here, \(A_o\) and \(A_g\) represent the attention maps for each patch from the original and generated images, respectively. These attention maps indicate the relative importance of different patches within each image.

\noindent The next step involves identifying and selecting patches based on their attention scores using the selection function \(\tau\):
\begin{equation}
P_{ro} = \tau(A_o, k)
\end{equation}
\begin{equation}
P_{rg} = \tau(A_g, k)
\end{equation}

\noindent where \(P_{ro}\) and \(P_{rg}\) represent the selected patches from the original and generated images, chosen based on having the highest attention scores, and \(k\) is a hyperparameter that represents the number of top patches to be selected.

\noindent \subsubsection{Patch-based Similarity Assessment}
The similarity between the obtained patches \(P_{ro}\) and \(P_{rg}\) is quantified using a modified version of the Dice coefficient, because we have now continuous image data, and Dice is usually used for binary data, so we modified Dice as:
\begin{equation}
D(P_{ro}, P_{rg}) = \frac{2 \times |P_{ro} \cap P_{rg}|}{|P_{ro}| + |P_{rg}|}
\label{eq:dice}
\end{equation}

\noindent where \(|P_{ro} \cap P_{rg}|\) represents the intersection of the patches, computed as the sum of the element-wise product, and \(|P_{ro}|\) and \(|P_{rg}|\) are the sums of the values in \(P_{ro}\) and \(P_{rg}\), respectively.

\noindent The overall Dice similarity for the images is then computed as an average of the Dice coefficients for all corresponding pairs of high-attention patches.
\begin{equation}
Dice = \frac{1}{N} \sum_{i=1}^{N} D(P_{ro_i}, P_{rg_i})
\label{eq:average_dice}
\end{equation}

where \(N\) is the number of patch pairs compared. Since we integrate Dice with \(S_2\) MMD where a lower score (close to 0) is better, so we also inverted Dice to align with this preference:
\begin{equation}
S_1 = 1 - Dice
\end{equation}
This inversion is essential as it aligns \(S_1\) with \(S_2\), both reflecting a preference for lower scores. 

\subsubsection{Deep Feature Extraction}
To extract deep features from these images, we utilize a last layer before the classification decision of a pre-trained Vision Transformer (ViT). 

\noindent The feature extraction of original image process can be expressed mathematically as follows:
\begin{equation}
f_o = \theta(\text{ViT}(I_o))
\label{eq:1}
\end{equation}
The deep features extraction for generated image can be expressing as:
\begin{equation}
f_g = \theta(\text{ViT}(I_g))
\label{eq:2}
\end{equation}

where $f_o$ and $f_g$ represent the deep features of $I_o$ and $I_g$, respectively. Here, $\theta$ denotes the transformation applied by the ViT before classification decision that condenses the spatial dimensions and returns deep features.

\subsubsection{Maximum Mean Discrepancy (MMD)}
In the previous section, we computed the similarity \(S_I\) using patches of images. However, in this section, we will examine the distributional differences in the entire image's deep features. For this purpose, we employed the Maximum Mean Discrepancy, which is a powerful non-parametric method used in statistics to compare the similarity between two distributions. It was first used in Kernel Inception Distance (KID), introduced by Binkowski et al. \cite{binkowski2018demystifying}. However, it was used only with the Polynomial Kernel and remains unexplored with other effective kernels. 

\noindent In the context of the GLIPS, we investigate the use of other kernels such as Radial Basis Function (RBF) and Exponential Kernels, which help evaluate how closely the feature distributions of the generated image align with those of the original image. Each kernel has hyperparameters that are optimized to measure the similarity between the deep feature distributions of the original ($I_o$) and generated ($I_g$) images.

Radial Basis Function (RBF) Kernel:
\begin{equation}
k_{\text{RBF}}(x, y) = \exp\left(-\gamma \|x - y\|^2\right)
\end{equation}
where $\gamma = \frac{1}{2\sigma^2}$ is the hyperparameter that determines the width of the Gaussian function.

Polynomial Kernel:
\begin{equation}
k_{\text{poly}}(x, y) = (\alpha x^\mathsf{T} y + c)^d
\end{equation}
where $\alpha$ is the slope, $c$ is the constant term, and $d$ is the degree of the polynomial.

Exponential Kernel:
\begin{equation}
k_{\text{exp}}(x, y) = \exp\left(-\frac{\|x - y\|}{\sigma}\right)
\end{equation}

where $\sigma$ determines the rate of the exponential decay. \(x\) and \(y\) are two points of original or generated deep feature distribution. Given the feature sets $f_o$ and $f_g$ derived from the real and generated images respectively, the MMD is calculated using a kernel function, a kernel function is selected using Hyperparameter optimization, typically Radial Basis Function (RBF), Polynomial or Exponential Kernel, to measure the distance between these sets in high-dimensional space:

\begin{align}
K_{oo} &= \frac{1}{N^2} \sum_{i,j=1}^{N}k(f_{o_i}, f_{o_j}) \\
K_{gg} &= \frac{1}{M^2} \sum_{i,j=1}^{M} k(f_{g_i}, f_{g_j}) \\
K_{og} &= \frac{2}{NM} \sum_{i,j=1}^{N,M} k(f_{o_i}, f_{g_j})
\end{align}

Then, the MMD is computed as:
\begin{equation}
S_2 = K_{oo} + K_{gg} - K_{og}
\label{eq:mmd}
\end{equation}

In the context of our dataset, where human perceptual scores directly constrain the image sets, the output range of our modified MMD inherently aligns with these scores, typically ranging between 0 and 1 due to the properties of the kernels and the normalization of the input features. This inherent alignment obviates the need for additional scaling of MMD scores.

Hence, our methodology benefits from maintaining the MMD in its native scale by preserving its direct interpretability and relevance to human-judged standards of image realism, so there is no scaling required and we can now integrate  \( S_2 \) with \( S_1 \)  by facilitating a balanced integration into the final GLIPS score where no single component disproportionately impacts the outcome.  MMD serves as a critical component in determining whether the generated images retain the statistical properties necessary to be considered photorealistic. By computing the MMD using a variety of kernel functions, we can assess the model's performance from different perspectives by offering a comprehensive understanding of the generative model's capabilities.

\begin{table*}[h]
\centering
\caption{Classification bins for Metrics}
\label{tab:imagbins}
\begin{tabular}{lcccccccc}
\toprule
\textbf{Criteria}          & \textbf{Score} & \textbf{SSIM}      & \textbf{PSNR}   & \textbf{FID}     & \textbf{MS-SSIM}  & \textbf{LPIPS}   & \textbf{GLIPS}     & \textbf{Inception Score} \\ \midrule
Strongly Disagree          & 0.0 - 1.0              & -1 to -0.6         & 0-7             & $>$ 150            & -1 to -0.6        & 0.9 to 1         & $>$ 0.8            & 0 $<$ 1                    \\ \midrule
Somewhat Disagree          & 1.1 - 2.0              & -0.5 to -0.2       & 8-15            & 100 to 149       & -0.5 to -0.2      & 0.7 to 0.8       & 0.7 to 0.8       & 1 $<$ 2                    \\ \midrule
Neutral                    & 2.1 - 3.0              & -0.1 to 0.2        & 16-23           & 31 to 99         & -0.1 to 0.2       & 0.5 to 0.6       & 0.5 to 0.6       & 2 $<$ 3                    \\ \midrule
Somewhat Agree             & 3.1 - 4.0              & 0.3 to 0.6         & 24-31           & 11 to 30         & 0.3 to 0.6        & 0.3 to 0.4       & 0.3 to 0.4       & 3 $<$ 5                    \\ \midrule
Strongly Agree             & 4.1 - 5.0              & 0.7 to 1.00        &  $>$ 32            &  $<$ 10             & 0.7 to 1.00       & 0 to 0.2         & 0 to 0.2         &  $>$ 6  \\ \bottomrule
\end{tabular}
\end{table*}
\subsubsection{Final Score}
The final score in the Global-Local Image Perceptual Score is computed by integrating the results from the patch-based continuous Dice similarity (S1) and the MMD assessment (S2). The GLIPS utilizes both the inverted Dice coefficient \(S_1\) and the MMD \(S_2\) to form a comprehensive metric that assesses image similarity on both micro and macro levels.  This combined metric provides a robust measure of image fidelity and realism, crucial for applications where precise image analysis is paramount.
The computation is as follows:

\begin{equation}
\text{S} = S_2 \times (1 - \lambda \times S_1)
\label{eq:final_score}
\end{equation}

where \( \lambda \) is a balancing parameter empirically selected to optimize the alignment between the combined score and human judgment scores. This formulation allows the integration of both local patch similarity and global distributional similarity, reflecting a comprehensive and balanced approach to evaluating image quality. The effectiveness of this scoring method has been validated through correlation analysis with human ratings, and the choice of \( \lambda \) was refined through sensitivity analysis to ensure robustness across different image sets.

\subsection{Interpolative Binning Scale (IBS)}
\label{subsec:quantitative_scaling}
To address the challenge of conducting a fair comparison between human and metric scores, we introduce the Interpolative Binning Scale (IBS), a novel quantitative scaling method. The IBS is designed to provide an interpretable understanding of metric scores by first classifying these values into predefined bins reflective of the score's interpretive meaning, and then refining these classifications through linear interpolation to find the exact position of the score within the bin. This two-step process allows for a high-resolution conversion of raw metric values into a scoring system that mirrors human evaluative gradations.

\subsubsection{Classification and Interpolation}
\label{subsubsec:class_interp}
In the classification step, raw scores from an image metric are discretely categorized into bins that represent levels of agreement or quality, such as 'Strongly Agree' to 'Strongly Disagree'. This qualitative mapping transforms the continuous spectrum of metric scores into a finite set of interpretable categories. The reason is that we often have human scores in a Likert scale ranging from 'Strongly Agree' to 'Strongly Disagree'. So converting metric scores into Likert bins provides a more interpretable situation.

The second step applies linear interpolation within these bins to assign a precise numerical score. Specifically, for a given value \( x \), and a set of ordered bins \( \{b_i\} \) with corresponding scores \( \{s_i\} \), the IBS score is calculated as follows:

\begin{equation}
\text{IBS Score} = s_i + \left(\frac{s_{i+1} - s_i}{b_{i+1} - b_i}\right) \times (x - b_i)
\label{eq:IBS}
\end{equation}

Here, \( s_{i+1} \) and \( s_i \) are the scores corresponding to the bins immediately above and below the value \( x \), and \( b_{i+1} - b_i \) are the boundaries of the bin within which \( x \) resides.
This basic method provides the fairest and most interpretable mechanism to conduct a comparison between human and metric scores.

\subsection{Illustrative Example}
\label{subsubsec:example_detailed}

In the context of image quality evaluation, let's take the example of SSIM, it is an essential metric that needs to be interpreted qualitatively. For the purpose of this example, let us consider an SSIM score of 0.45. We aim to scale this score using the proposed Interpolative Binning Scale method.

\noindent \subsubsection{Classification}
Initially, the SSIM score is classified into one of the predefined categories that represent various levels of image quality. These categories, or bins, are detailed in Table \ref{tab:imagbins}.

The SSIM score of 0.45 is categorized into the 'Somewhat Agree' bin.

\noindent \subsubsection{Interpolation}
The next step involves interpolating the score within this bin. The formula used for interpolation within the 'Somewhat Agree' range is as follows:

\begin{equation}
\text{IBS Score} = 3.1 + \left(\frac{5 - 4}{0.6 - 0.3}\right) \times (0.45 - 0.3)
\label{eq:IBS_example}
\end{equation}

Calculating the above formula provides us with the interpolated IBS score:

\begin{equation}
\text{IBS Score} = 3.1 + \left(\frac{1}{0.3}\right) \times (0.15) = 4 + 0.5 = 3.6
\label{eq:IBS_calculation}
\end{equation}

\paragraph{Interpretation}
The IBS score of 3.6 indicates that the original SSIM value of 0.45 aligns closer to the 'Somewhat Agree' end of the score spectrum. This score is particularly meaningful as it provides a refined and precise placement of the SSIM value within its bin. Moreover, this scoring method is superior to traditional normalization techniques such as Z-Standardization and Min-Max scaling, as well as linear regression-based black-box methods, as it mitigates their inherent biases and offers a more interpretable score that is understandable to humans.

\section{Evaluation and Results}
This section will provide details on the selection of Image Generative Models, the Human Study, and the results of the experiments we conducted. \label{sec:results}

\begin{table*}[!h]
\centering
\caption{Photoelectric Image Quality Scores of Different Models Rated by Human}
\label{tab:image_quality}
\begin{tabular}{@{}lccccc@{}}
\toprule
Question Text & Camera Generated & DALL.E2 & Glide & Stable Diffusion & DALL.E3 \\ \midrule
The image looks like a photograph of a real scene. & 4.12 & 3.61 & 2.05 & 3.27 & 2.66 \\
I can easily imagine seeing this image in the real world. & 4.30 & 3.80 & 2.04 & 3.34 & 2.84 \\
The visual details in this image make it appear realistic. & 4.10 & 3.56 & 1.94 & 3.32 & 2.76 \\
The textures in the image look natural and real. & 3.95 & 3.47 & 2.03 & 3.23 & 2.61 \\
The lighting and shadows in the image contribute to its realism. & 3.83 & 3.70 & 2.15 & 3.35 & 2.87 \\
\midrule
Average Response (Out of 5) & 4.06 & 3.63 & 2.04 & 3.30 & 2.75 \\ \bottomrule
\end{tabular}
\end{table*}

\subsection{Human Study}

\begin{figure}[!t]
    \centering
    \includegraphics[width=0.5\textwidth]{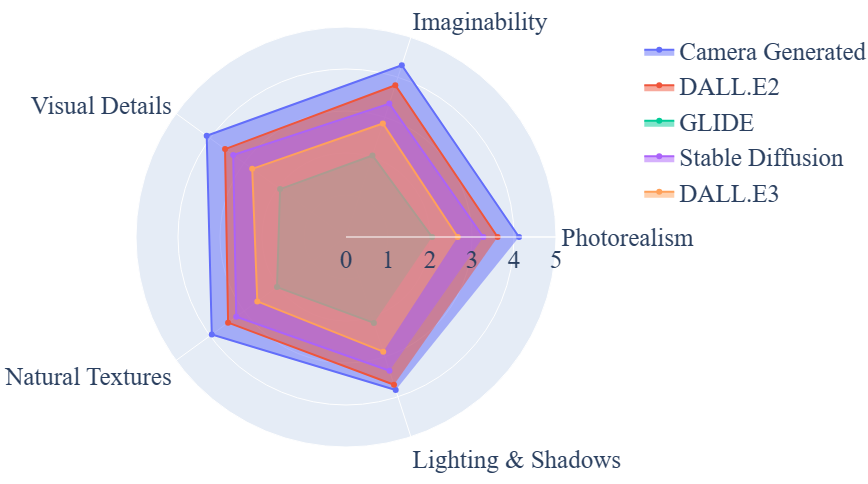}
    \caption{Comparative Analysis of Photorealistic Image Quality Across Different AI Models}
    \label{fig:survey}
\end{figure}

We designed and conducted a human study divided into three parts: a) Photorealism, b) Image Quality, and c) Text to Image Alignment, each comprising five questions. This paper focuses solely on the Photorealism component, which assesses human perceptions of photorealistic quality. The questions in the Photorealism section were carefully selected to cover the core aspects of realism: Does the image look like a real scene? Could the image exist in the real world? How authentic are the visual details? How natural are the textures? How believable are the lighting and shadows? 

\noindent Our study's ethics approval was granted by the Western University Non-Medical Research Ethics Board (NMREB), Canada, under project ID: 124753.

For the study, we presented both original and AI-generated images to 350 participants recruited through the Prolific platform \cite{prolific2024}. We selected original images from the widely recognized COCO (Common Objects in Context) dataset \cite{cocodataset}, which comprises 330,000 images and includes human-written captions for each image. We selected 20 images from this dataset for our study to compare with our proposed metric.

Participants were also shown images generated by four prominent image generative models: Stable Diffusion \cite{lee2023diffusion}, DALLE-2 \cite{marcus2022preliminary}, Glide \cite{nichol2021glide}, and DALL-E3 \cite{betker2023improving}. This selection, which includes a blend of technologies such as Transformers, Variational Autoencoders, and Diffusion models, represents both industry and academic research outputs. This diverse range will help investigate the effectiveness of different technologies in generating photorealistic images, as depicted in Figure \ref{fig:image}. 

Participants rated both the camera-captured and AI-generated images. The camera-generated images scored an average of 4.06 out of 5.00, while DALLE-2 scored 3.63, Glide 2.04, Stable Diffusion 3.30, and DALL-E3 2.75 on the photorealism scale. Detailed results are presented in Table \ref{tab:image_quality}. We presented a graphical illustration of the results in Figure \ref{fig:survey}. The publicly available link to the human scores and demographics can be found at: \href{https://github.com/udanish50/VisualVerity}{https://github.com/udanish50/VisualVerity}.

\subsection{Selection of Metrics for Comparisons}
We have chosen to compare our proposed method against several established metrics for assessing photorealism, namely FID, LPIPS, KID, SSIM, and MS-SSIM. The rationale behind this selection is twofold: first, we included baseline metrics such as FID and KID, which are commonly used deep features-based metrics to evaluate the photorealism of generated images. Second, LPIPS offers a similar measure to FID and KID but utilizes activations of pre-trained models rather than deep features and provides a perceptual sense of images. Additionally, we employed two pixel-based metrics, SSIM and its multi-scale version MS-SSIM, to ensure a comprehensive comparison with our proposed metric. This selection ensures a thorough evaluation of our proposed method in relation to existing approaches. 

To compare the human and metric scores numerically, we employed two statistical measures: Mean Absolute Difference (MAD) and Mean Absolute Percentage Error (MAPE). These can be defined as follows:

\begin{equation}
\text{MAD} = \frac{1}{N} \sum_{i=1}^{N} |x_i - y_i|
\label{eq:MAD}
\end{equation}

\begin{equation}
\text{MAPE} = \frac{100\%}{N} \sum_{i=1}^{N} \left|\frac{x_i - y_i}{x_i}\right|
\label{eq:MAPE}
\end{equation}

where \( N \) is the total number of observations, \( x_i \) represents the human scores, and \( y_i \) denotes the metric scores for the \( i \)-th observation.

\begin{table*}[h!]
\centering
\caption{Comparison of Metric Scores with Human Judgment on Photorealism}
\label{tab:metric_comparison}
\begin{tabular}{|p{2.3cm}|p{1.85cm}|p{0.9cm}|p{0.9cm}|p{0.9cm}|p{2cm}|p{2cm}|p{0.9cm}|p{0.9cm}|}
\toprule
\textbf{Model} & \textbf{Metric} & \textbf{Actual Score} & \textbf{Rescaled Score} & \textbf{Human Score} & \textbf{Likert Category (Metric)} & \textbf{Likert Category (Human)} & \textbf{MAD} & \textbf{MAPE} \\ \midrule
Stable Diffusion & \multirow{4}{*}{FID} & 28.83 & 4.52 & 3.30 & Strongly Agree & Somewhat Agree & 1.22 & 36.96 \\
\cmidrule(r){1-1} \cmidrule(lr){3-9}
DALLE2 & & 13.81 & 4.90 & 3.63 & Strongly Agree   & Somewhat Agree & 1.27 & 34.98 \\
\cmidrule(r){1-1} \cmidrule(lr){3-9}
GLIDE & & 21.28 & 4.71 & 2.04 & Strongly Agree & Neutral & 2.67 & 130.80 \\
\cmidrule(r){1-1} \cmidrule(lr){3-9}
DALLE3 & & 41.81 & 4.20 & 2.75 & Strongly Agree & Neutral & 1.45 & 52.72 \\
\midrule
Stable Diffusion & \multirow{4}{*}{LPIPS} & 0.78 & 2.05 & 3.30 & Neutral & Somewhat Agree & 1.25 & 37.87 \\ \cmidrule(r){1-1} \cmidrule(lr){3-9}
DALLE2 &  & 0.66 & 2.67 & 3.63 & Neutral & Somewhat Agree & 0.96 & 26.44 \\ \cmidrule(r){1-1} \cmidrule(lr){3-9}
GLIDE &  & 0.67 & 2.65 & 2.04 & Neutral & Neutral & 0.60 & 29.90 \\ \cmidrule(r){1-1} \cmidrule(lr){3-9}
DALLE3 &  & 0.74 & 2.29 & 2.75 & Neutral & Neutral & 0.45 & 16.72 \\ \midrule
Stable Diffusion & \multirow{4}{*}{SSIM} & 0.153 & 3.88 & 3.30 & Somewhat Agree & Somewhat Agree & 0.58 & 17.57 \\ \cmidrule(r){1-1} \cmidrule(lr){3-9}
DALLE2 &  & 0.195 & 3.98 & 3.63 & Somewhat Agree & Somewhat Agree  & 0.35 & 9.64 \\ \cmidrule(r){1-1} \cmidrule(lr){3-9}
GLIDE &  & 0.396 & 4.49 & 2.04 & Strongly Agree & Neutral & 2.45 & 120.09 \\ \cmidrule(r){1-1} \cmidrule(lr){3-9}
DALLE3 &  & 0.115 & 3.78 & 2.75 & Somewhat Agree & Neutral & 1.02 & 37.45 \\ \midrule
Stable Diffusion & \multirow{4}{*}{MS-SSIM} & 0.128 & 3.82 & 3.30 & Somewhat Agree & Somewhat Agree & 0.52 & 15.75 \\ \cmidrule(r){1-1} \cmidrule(lr){3-9}
DALLE2 &  & 0.134 & 3.83 & 3.63 &  Somewhat Agree & Somewhat Agree  & 0.20 & 5.50 \\ \cmidrule(r){1-1} \cmidrule(lr){3-9}
GLIDE &  & 0.139 & 3.84 & 2.04 & Somewhat Agree & Neutral & 1.79 & 88.23 \\ \cmidrule(r){1-1} \cmidrule(lr){3-9}
DALLE3 &  & 0.058 & 3.64 & 2.75 & Somewhat Agree & Neutral & 0.89 & 32.36 \\ \midrule

Stable Diffusion & \multirow{4}{*}{KID} & 0.98 & 1 & 3.30 & Strongly Disagree & Somewhat Agree & 2.30 & 69.69 \\ \cmidrule(r){1-1} \cmidrule(lr){3-9}
DALLE2 &  & 0.67 & 1.15 & 3.63 & Somewhat Disagree & Somewhat Agree & 2.48 & 68.31 \\ \cmidrule(r){1-1} \cmidrule(lr){3-9}
GLIDE &  & 0.33 & 3.07 & 2.04 & Somewhat Agree & Neutral & 1.02 & 50.49 \\ \cmidrule(r){1-1} \cmidrule(lr){3-9}
DALLE3 &  & 0.66 & 1.16 & 2.75 & Somewhat Disagree & Neutral & 1.59 & 57.81 \\ \midrule

Stable Diffusion & \multirow{4}{*}{GLIPS \(\lambda = 0.54\)} & 0.37 & 3.12 & 3.30 & Somewhat Agree & Somewhat Agree & 0.17 & 5.45 \\ \cmidrule(r){1-1} \cmidrule(lr){3-9}
DALLE2 &  & 0.32 & 3.82 & 3.63 & Somewhat Agree & Somewhat Agree & 0.19 & 5.23 \\ \cmidrule(r){1-1} \cmidrule(lr){3-9}
GLIDE &  & 0.73 & 1.34 & 2.04 & Somewhat Disagree & Neutral & 0.70 & 34.31 \\ \cmidrule(r){1-1} \cmidrule(lr){3-9}
DALLE3 &  & 0.55 & 2.20 & 2.75 & Neutral & Neutral & 0.54 & 
19.99 \\ \midrule 

Stable Diffusion & \multirow{4}{*}{GLIPS \(\lambda = 0.62\)} & 0.34 & 3.28 & 3.30 & Somewhat Agree & Somewhat Agree & 0.02 & 0.61 \\ \cmidrule(r){1-1} \cmidrule(lr){3-9}
DALLE2 &  & 0.23 & 3.35 & 3.63 & Somewhat Agree & Somewhat Agree & 0.28 & 7.71 \\ \cmidrule(r){1-1} \cmidrule(lr){3-9}
GLIDE &  & 0.62 & 1.85 & 2.04 & Somewhat Disagree & Neutral & 0.18 & 8.95 \\ \cmidrule(r){1-1} \cmidrule(lr){3-9}
DALLE3 &  & 0.48 & 2.56 & 2.75 & Neutral & Neutral & 0.17 & 
6.23 \\  \bottomrule
\end{tabular}
\end{table*}

\subsection{Classification Bins for Metrics}
To implement our proposed scaling strategy, we designed bins classified against a Likert scale ranging from "strongly disagree" to "strongly agree," with scores ranging from 1 to 5. As mentioned earlier, these bins were created based on the results of a human study conducted by experts from the Human-Centric Computing Group at Western University Canada. Due to the complexity of the task and the need for expertise in computer vision and AI-human interaction, an open survey was not feasible, necessitating a study conducted by specialized experts in the field. The classification bins can be found in Table \ref{tab:imagbins}. It is worth noting that KID also utilizes similar bins to our proposed metric.
\begin{figure*}[h!]
    \centering
    \includegraphics[width=0.9\textwidth]{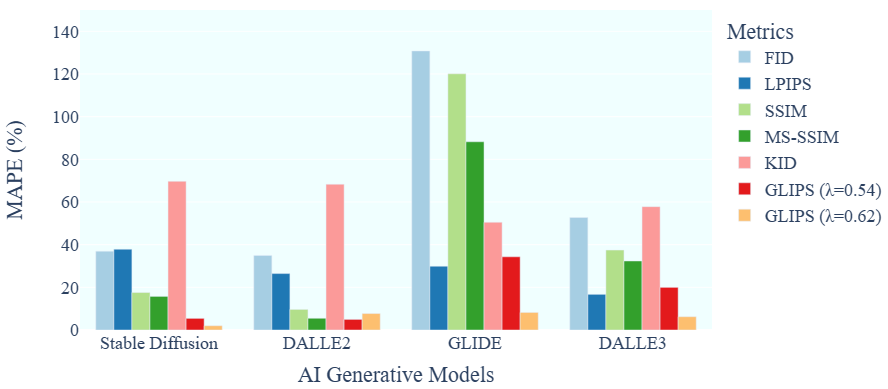} 
    \vspace{-13pt}
    \caption{Comparison of GLIPS with other approaches for each model.}
    \label{fig:lineoverall}
\end{figure*}

%\subsection{Hyper-Parameter Tuning}
%In the GLIPS, we recognize the importance of selecting hyperparameters such as \(k\), which determines the number of significant patches analyzed, and parameters \(\sigma\), \(\alpha\), and \(\gamma\) associated with kernel functions. Unlike traditional hyperparameter optimization techniques like GridSearch, which are unsuitable due to GLIPS being a metric and not a model, we employ an empirical approach. We empirically selected various kernel functions whether RBF, Polynomial, or Laplacian—significantly affects the computation of similarities and is thus considered a critical hyperparameter. Each kernel type is selected based on its ability to enhance the metric's sensitivity to photorealistic image quality specific to various generative models. Through empirical evaluation, we fine-tune these parameters, ensuring that GLIPS robustly captures both local patch similarities and global distributional characteristics. We report most results on most optimal hyperparameters, such as RBF kernel is most suited in all cases, with its gamma value of 4, but just on one case of GLIDE model, \(\gamma\) value was 10. While we selected mostly the top 35 image patches where attention scores are high.  

\subsection{Results and Analysis}
The Table \ref{tab:metric_comparison} provides complete details and interoperable insight into results, "Actual Score" in Table \ref{tab:metric_comparison}  refers to the score obtained directly from the metric, whereas "Rescaled Score" indicates that the score has been adjusted using the our proposed IBS technique. The Likert category of the metric provides a direct and understandable comparison in a subjective sense, whereas the MAD quantifies the absolute difference between the human and metric scores. The MAPE expresses the error in terms of a percentage.

The FID shows a substantial deviation from human judgment across various models. Specifically, the Mean Absolute Percentage Error (MAPE) for FID stands at 36.96\% for Stable Diffusion, 34.98\% for DALLE2, sharply increases to 130.30\% for GLIDE, and remains at 52.72\% for DALLE3. These figures underscore that FID scores deviate significantly—by at least 30\%—from human scores. Furthermore, a comparative analysis of Likert categories reveals that FID consistently fails to align with human-assessed categories; where humans tended to rate images from Stable Diffusion and DALLE2 as 'Somewhat Agree', and GLIDE and DALLE3 as 'Neutral'. This inconsistency highlights FID's inability to capture the essence of image quality as perceived by humans.

The LPIPS metric shows comparatively better alignment with human judgment than the FID. Specifically, LPIPS achieves a MAPE of 37.87\% for Stable Diffusion, an improvement over its FID counterpart. For DALLE2, the error is reduced to 26.44\%, and for GLIDE, the error decreases to 29.90\%, which is nearly a 100\% improvement compared to FID's performance. Most notably, LPIPS records a minimal error of 16.72\% for DALLE3, the lowest among all metrics evaluated for this model. This metric matches the human judgment of 'Neutral' for both GLIDE and DALLE3, illustrating its superior capability in capturing nuances more reflective of human perception than FID.

The SSIM shows a promising correlation with human perception. For instance, the SSIM for Stable Diffusion showed a rescaled score of 3.88 closely aligning with the human score of 3.30, which reflects a high degree of perceptual alignment as both fall into the 'Somewhat Agree' category on the Likert scale with just 17.17\% error. However, notable discrepancies emerge in cases such as GLIDE, where despite a high SSIM rescaled score of 4.49 indicating 'Strongly Agree', the corresponding human score was only 2.04 'Neutral', resulting in a substantial MAPE of 120.09\%. This significant variance in MAPE highlights the occasional challenges SSIM faces in mirroring human judgments across different settings and suggesting areas for further refinement.

The MS-SSIM offers an extended analysis of image quality by assessing visual details across multiple scales. As shown in Table \ref{tab:metric_comparison}, the alignment between the MS-SSIM scores and human judgments varies but it gives impressive improvement to SSIM alone. For instance, in the case of Stable Diffusion, the MS-SSIM score closely mirrors the human score 3.82 vs. 3.30, and both are categorized as 'Somewhat Agree', indicating a consistent perception of image quality. This scenario exhibits a relatively low MAPE of 15.75\%, suggesting a good match between the metric and human perception. Similarly, there is a low error in the case of DALLE2 of just 5.50\%. However, GLIDE shows a divergence; despite an MS-SSIM score indicating 'Somewhat Agree' 3.84, the human judgment falls under 'Neutral' (2.04), resulting in a significantly higher MAPE of 88.23\%.

We implemented KID with three different kernels namely Polynomial, RBF, and Exponential. The KID metric shows varied accuracy when compared to human judgments. For instance, in the case of Stable Diffusion, a misalignment is evident with a high MAPE of 69.69\%, where the metric categorizes image quality as 'Strongly Disagree', starkly contrasting the human judgment categorized under 'Somewhat Agree'. Similarly, for DALLE2, the discrepancy remains significant with a MAPE of 68.31\%, again reflecting a serious mismatch in the perceptual evaluation as the metric suggests 'Somewhat Disagree' versus the human-assigned 'Somewhat Agree'. The model GLIDE shows a MAPE of 50.49\%, where KID slightly improves in capturing human perception aligning 'Somewhat Agree' with a 'Neutral' human rating. The smallest MAPE observed is with DALLE3 at 57.81\%, where both metric and human judgment categorize the image quality under a neutral standpoint.

We compared our proposed model GLIPS at different $\lambda$ values, after conducting sensitivity analysis, we presented results on two different values where results were good at 0.54 and 0.62. At the lambda value, 0.54 metric shows impressive performance in mirroring human scores of photorealistic image quality, For instance, in the case of Stable Diffusion, GLIPS indicates a MAPE of just 5.45\%, which shows a high level of agreement with human, categorizing both metrically and human-rated as 'Somewhat Agree'. For DALLE2, the MAPE slightly increases to 5.23\% but maintains consistent alignment in the 'Somewhat Agree' category. Moving to the GLIDE model, GLIPS showcases a higher MAPE of 34.31\%, with the metric score 'Somewhat Disagree' and human aligning to 'Neutral'. This indicates a divergence in perception assessed by GLIPS compared to human. The analysis of DALLE3 presents a MAPE of 19.99\%, the highest among the models. Here, both GLIPS and human judgments correspond to a 'Neutral' category by reflecting a consistency in assessment. 

Next, at $\lambda$ = 0.62, GLIPS not only outperformed all other metrics but also reduced the MAPE to less than 10\% across all models tested. For Stable Diffusion, GLIPS achieved an impressive MAPE of just 0.61\%, closely matching human assessment. For DALLE2, the MAPE remained at 7.71\%. Notably, in the case of GLIDE, where other metrics struggled, GLIPS significantly reduced the error to 8.95\%, nearing human judgment levels. For DALLE3, GLIPS reduced the MAPE to 6.23\%. In all instances, the Likert classification matched human ratings, except in the case of GLIDE, where GLIPS stood at 'Somewhat Disagree' and human perception at 'Neutral', with only a 0.18 mean absolute deviation. These results affirm our initial hypothesis: GLIPS not only reduces error close to human assessments by evaluating patch-level local similarity and global distributional characteristics but also ensures no critical information from the image is missed, thus providing a robust approach to assessing the photorealistic quality of images.

We also conducted an ablation study to evaluate the effectiveness of $S_1$ equation: \ref{eq:average_dice} local patch similarity and $S_2$ equation: \ref{eq:mmd} global distributional similarity separately, to justify the necessity of combining them to reduce the deviation from human scores. For $S_1$ alone, the ability to mirror human judgment is apparent. Consider the first example with Stable Diffusion, where $S_1$ yields a score of 3.80 and $S_2$ produces a score of 3.01. It is important to note that both scores have been rescaled using our proposed scaling method IBS. In the case of Stable Diffusion, the human score is 3.30. It can be clearly observed that neither $S_1$ nor $S_2$ alone can closely match the human score. However, by combining both at a specified $\lambda$, we achieve a score of 3.28, which is very close to the human score, with an error of just 1.99\%. This underscores the effectiveness of our proposed GLIPS and the essential need to consider both local and global distributional similarities of images.

\section{Conclusion}
\label{sec:conclusion}
This study introduced and validated the Global-Local Image Perceptual Score is a novel metric for assessing the photorealistic image quality of AI-generated images. By integrating patched-based attention mechanisms with Maximum Mean Discrepancy of entire image features, GLIPS provides a robust measure that significantly correlates with human visual judgments,  outperforming traditional and commonly used metrics across several benchmarks. The GLIPS methodologically advances the field of photorealistic image quality evaluation by addressing critical flaws inherent in existing metrics which are generally tailored for camera-generated images. Traditional metrics such as FID, SSIM, and their derivatives often fail to capture the full spectrum of human perception, as evidenced by their substantial MAPE in comparative tests. In contrast, GLIPS demonstrates a consistently lower MAPE by indicating a higher alignment with human assessments, particularly for images from leading-edge models like Stable Diffusion and DALLE2.  

Furthermore, the development of the Interpolative Binning Scale (IBS) for scaling metric scores to align with human judgment has shown to reduce bias significantly and provide human interpretable results that can be trusted. This method refines the resolution of categorization within the GLIPS framework by ensuring that the metric's sensitivity to the image is accurately reflected in its scoring. The implications of this research are manifold. For one, by providing a reliable tool that mirrors human perception, GLIPS can be instrumental in the iterative development of generative models, potentially guiding enhancements in photorealism and detail. Moreover, its application extends beyond mere academic interest to practical use-cases in media production, medical imaging, and other fields where photorealistic image quality is critical.

Future work will aim to further optimize the GLIPS by exploring different configurations of the underlying neural network architectures and refining the kernel functions used in the MMD calculation. These enhancements will seek to broaden the metric's applicability to a wider range of image types and generative models by ensuring that GLIPS remains effective in the face of evolving AI capabilities. In sum, the GLIPS marks a significant step forward in the objective evaluation of AI-generated images. It aligns closely with human visual standards, thereby providing a robust tool for developers and researchers aiming to improve the fidelity and realism of synthetic imagery.
\bibliography{biblography.bib}
\end{document}